\documentclass[letterpaper, 10 pt, conference]{ieeeconf}  
\IEEEoverridecommandlockouts     
\let\labelindent\relax                      
 
\usepackage[algo2e,ruled,linesnumbered,vlined,boxed,commentsnumbered]{algorithm2e}
\usepackage{amsmath}
\usepackage{multirow}
\usepackage{array}
\usepackage{graphicx}
\usepackage{cite}
\usepackage{enumitem}
\usepackage{algorithmic}
\usepackage{hyperref}
\usepackage{caption}
\usepackage{float} 
\usepackage{subcaption}
\usepackage{threeparttable}
\usepackage{bm}
\usepackage{amssymb}

\usepackage{flushend} 

\usepackage{caption}
\makeatletter
\newcommand{\resetAlgoLine}{%
    \setcounter{AlgoLine}{0}%
}
\makeatother
\let\oldalgorithm=\algorithm
\def\algorithm{%
    \resetAlgoLine 
    \oldalgorithm  
}

\title{\LARGE \bf
RRT*former: Environment-Aware Sampling-Based Motion Planning using Transformer
}

\author{Mingyang Feng, Shaoyuan Li and Xiang Yin
\thanks{This work was supported by the National Natural Science Foundation of China (62173226, 92367203). 
 }
\thanks{M. Feng, S. Li and X. Yin  are  with the School of Automation and Intelligent Sensing, Shanghai Jiao Tong University, Shanghai 200240, China. (Corresponding Author: Xiang Yin) {\tt  E-mail: \{Fmy-135214,syli,yinxiang\}@sjtu.edu.cn}.}  
}

\begin{document}

\maketitle
\thispagestyle{empty}
\pagestyle{empty}

\begin{abstract}
We investigate the sampling-based optimal path planning problem for robotics in complex and dynamic environments. Most existing sampling-based algorithms neglect environmental information or the information from previous samples. Yet, these pieces of information are highly informative, as leveraging them can provide better heuristics when sampling the next state. In this paper, we propose a novel sampling-based planning algorithm, called \emph{RRT*former}, which integrates the standard RRT*  algorithm with a Transformer network in a novel way. Specifically, the Transformer is used to extract features from the environment and leverage information from previous samples to better guide the sampling process.  
Our extensive experiments demonstrate that, compared to existing sampling-based approaches such as RRT*, Neural RRT*, and their variants, our algorithm achieves considerable improvements in both the optimality of the path and sampling efficiency. The code for our implementation is available on \url{https://github.com/fengmingyang666/RRTformer}.
\end{abstract}

\section{INTRODUCTION}
Motion planning in the presence of obstacles is a fundamental challenge in robotics and autonomous systems. The primary goal of this problem is to  navigate a robot from a starting point to a goal point while avoiding collisions with obstacles in its environment \cite{yin2024formal,zhao2025no}. Traditional search-based approaches to this problem, such as the A* and D* algorithms \cite{stentz1995focussed}, have been widely used due to their conceptual simplicity and efficient performance in low-dimensional spaces. However, these methods often struggle with computational complexity and scalability when applied to real-time motion planning in high-dimensional environments. As a result, there is a growing need for more advanced techniques that can handle the increased complexity and dynamic nature of modern robotic applications.

Sampling-based methods, such as Probabilistic Roadmaps (PRM), Rapidly-exploring Random Trees (RRT), and RRT* \cite{kavraki1996probabilistic,lavalle2006planning,karaman2011sampling}, have emerged as powerful alternatives due to their ability to handle high-dimensional spaces and complex obstacle geometries. These algorithms work by randomly sampling points in the configuration space and incrementally building a graph or tree structure that connects the start and goal points. Over the past years, sampling-based motion planning has been successfully applied in various complex engineering systems, including unmanned aerial vehicles \cite{butler2024sampling}, autonomous driving \cite{wang2022gmr} and robot manipulators \cite{huh2018constrained}.  
More recently, sampling-based approaches have also been extended to handle more complex planning tasks, such as those involving high-level specifications expressed in temporal logic \cite{kantaros2019sampling,yu2022security,vasile2020reactive,kantaros2020stylus,luo2021abstraction,liu2025zero, cui2024robust,li2023temporal}.

A core aspect of sampling-based planning is the  exploration strategy  used to generate samples. Traditional methods often rely on uniform random sampling, which can be inefficient in environments with complex obstacle structures or narrow passages. To improve sample efficiency, many approaches have been developed to better guide the sampling process toward regions that are more likely to yield optimal solutions.  
For instance,  Informed RRT  \cite{gammell2014informed} uses an admissible ellipsoidal heuristic to focus sampling within a subset of the configuration space that is guaranteed to contain better solutions. In \cite{qureshi2016potential,tahir2018potentially}, artificial potential fields are employed to guide samples toward the goal while avoiding obstacles. However, these methods typically rely on hand-crafted heuristics, which can be challenging to adapt to new or highly complex environments. 

More recently, learning-based approaches have been developed to enhance the sample efficiency in optimal path planning in complex environments. 
For example, in \cite{li2018neural}, a simple multi-layer perceptron (MLP) is used to estimate the cost of the path, attempting to solve the planning problem with nonlinear kinematic constraints. 
In \cite{wang2020neural}, convolutional neural networks (CNN) is used for environmental processing to obtain a smaller sampling area. 
In \cite{liu2024kg,liu2024nngtl}, graph neural networks (GNNs) are used to model the relationships between nodes to prevent collisions. In \cite{huang2024neural}, the authors use point cloud information to represent the sampling area, and use pointnet to further refine the possible sampling area. 
In \cite{chaplot2021differentiable}, using the entire map as input, a transformer is used to predict the cost of all pixels to the target point, and this is used for path planning. 
In \cite{johnson2021motion}, the transformer is used to encode environmental information and classify pixels to obtain sampling areas. 

A common feature of the above methods is that they focus on environmental information and only adjust the sampling area before sampling begins, without iteratively updating it as sampling proceeds. Recently, there have been efforts to leverage information from previous sampling points to better guide future samples.  
For example, in \cite{qureshi2020motion}, environmental information and previous sampling points are fused and input into a MLP to predict the next sampling point. However, this method struggles with processing complex long-distance paths because MLPs are not well-suited for handling long sequences.  
In \cite{chen2019learning}, reinforcement learning algorithms are used to continuously iterate and interact the sampling and learning processes, enabling the system to learn promising exploration directions based on the environment's structure. However, it relies on grid-based encoding of the entire workspace as input and requires the system to perceive the environment through learning rather than directly accessing environmental information.  
In \cite{johnson2023learning}, path points from the training set are encoded into discrete vectors using a transformer. 
The method infers the possible distribution area of each sampling point in a new environment by matching these vectors with environmental features. However, it only adjusts the sampling distribution and does not precisely determine the sampling points.

In this paper, we propose a new approach to improve sample efficiency by leveraging both environmental information and previous samples. The overall algorithm, called RRT*former, integrates the standard RRT* algorithm with the Transformer network architecture \cite{vaswani2017attention}. Specifically, environmental feature information is incorporated as positional encodings, enabling the algorithm to better capture the relationship between the sampling process and the environment. Sampling nodes serve as both the input and output of the Transformer, ensuring consistency between input and output. This design allows previous sampling nodes to be utilized as historical information in subsequent iterations, facilitating a better understanding of the relationships between sampling nodes. Since each sampling step is based on the current map and new information is continuously incorporated, the network can iteratively update its sampling strategy. This iterative update process is particularly crucial in dynamic environments, where conditions and information may change over time.  
We compare the proposed RRT*former approach with the standard RRT* algorithm and other neural network-guided sampling algorithms, such as NRRT*\cite{wang2020neural} and NIRRT*\cite{huang2024neural}. Experimental results demonstrate that our approach effectively reduces sampling time and the number of sampling nodes.

The rest of the paper is organized as follows. In Section II, we first formulate the path planning problem. Then, in Section III, we present our main algorithm, RRT*former, including its architecture and training details. In Section IV, extensive numerical experiments and simulations are provided to illustrate the effectiveness of our approach. Finally, we conclude the paper in Section V.

\section{Problem Formulation}

Let \( \mathcal{X} \) denote the configuration space, where \( \mathcal{X}_{obs} \) represents the obstacle space, and \( \mathcal{X}_{free} = \mathcal{X} \setminus \mathcal{X}_{obs} \) defines the free space that is navigable by the robot. Given a start state \( x_s \), a goal state \( x_g \), and a specific environment \( Env \), the goal is to find a feasible path \( \mathcal{P} \) that satisfies the following conditions:

\begin{itemize}
     \item
    $\mathcal{P}=\{x_i\in \mathcal{X}\}_{i=0}^{n}$
    \item 
    $x_i \in \mathcal{X}_{free},$  for all $x_i\in \mathcal{P}$
    \item 
    $Line(x_i,x_{i+1}) \in \mathcal{X}_{free},$  for all $x_i\in \mathcal{P}$
    \item 
    $x_0=x_s, x_n=x_g $
\end{itemize}

To quantify the quality of the path, we define its length using the Euclidean distance between consecutive nodes. The total path cost is given by
\[
\text{Cost}_{\mathcal{P}} = \sum_{i=1}^{n} \sqrt{\sum_{j=1}^{d} \left( x_i^{(j)} - x_{i-1}^{(j)} \right)^2},
\]
where \( n \) is the total number of nodes in the path and \( d \) is the spatial dimension of the configuration space \( \mathcal{X} \). For a 2D environment, \( \mathcal{X} \in \mathbb{R}^2 \); for a 3D environment, \( \mathcal{X} \in \mathbb{R}^3 \). This cost function measures the cumulative Euclidean distance between consecutive nodes, which is typically used to evaluate the length of the path.

At each iteration step \( t \), the configuration space is updated to reflect changes in the environment, particularly the obstacles. This update is expressed as:
\[
\Delta \mathcal{X}_{obs} = f(Env(t), t, x_t),
\]
where \( f \) is a sensor-based function that takes the current environmental data \( Env(t) \), the current time step \( t \), and the robot’s current configuration \( x_t \) as inputs. In static environments, where obstacles do not change over time, no update occurs, i.e., \( \Delta \mathcal{X}_{obs} = 0 \). In dynamic environments, the function \( f \) evolves as the environment changes.

Let \( \mathcal{X}_{tree}(t) = \{ x_i \mid x_i \in \mathcal{X}, i < t \} \) denote the set of previously sampled nodes up to time \( t \). Based on the principles of sample-based algorithms, \( \mathcal{X}_{tree} \) is constructed incrementally by sampling nodes from the configuration space. The process of adding new samples to \( \mathcal{X}_{tree} \) is influenced by the obstacle information at each time step. The quality of the samples generated during this process directly impacts the algorithm's convergence rate and the path's length. Therefore, it is crucial to design a sampling strategy that efficiently avoids obstacles. That is, we need to design a \textit{Sampler} that generates the next node \( x_{next} \) based on both the current environment information and the past samples in \( \mathcal{X}_{tree}(t) \). Specifically, the sampler function is expressed as:
\[
x_{next} = \text{Sampler}(f, \mathcal{X}_{tree}(t)),
\]
which guides the planning algorithm to explore the configuration space within the constraints of the environment.

\section{RRT*former Algorithm}
In this section, we present our main sampling-based algorithm, RRT*former. Specifically, we use the standard RRT* as the basic skeleton for trajectory sampling. When sampling the next state in the RRT* algorithm, we employ a transformer architecture to generate the sampling point, leveraging both environmental information and previous sampling data. Finally, we provide a detailed description of how the proposed neural network is trained.

\begin{figure*}[t]
\centering
\includegraphics[scale=0.116]{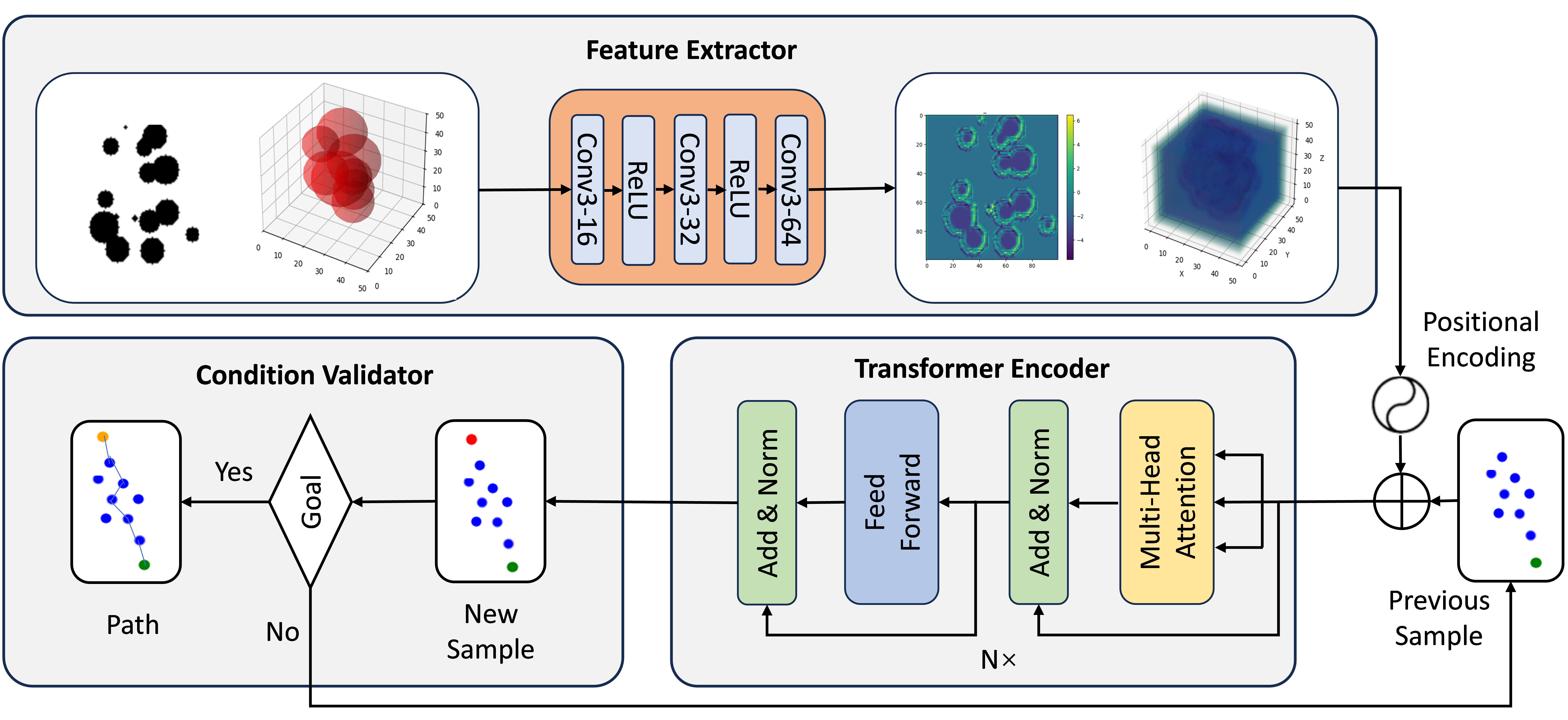}
\caption{Sampler Model. The Sampler consists of three parts. \textbf{Feature Extractor}: extract features from the environment using CNN. \textbf{Transformer Encoder}: generate new sample from previous sampling information and environment features. \textbf{Condition Validator}: determine when to stop sampling by checking whether the new sample is close enough to the goal.}
\label{Sampler_model}
\end{figure*} 

\subsection{The Basic RRT* Algorithm}
In this work, we adopt RRT* as the base sampling algorithm. The idea of RRT* is similar to the standard RRT algorithm. Specifically, in RRT*, the tree, denoted as \(\mathcal{X}_{tree}\), is iteratively grown by adding nodes that not only extend toward random samples in the configuration space \(\mathcal{X}\) but also aim to improve the path from the starting node to the goal.  

At each iteration, the following steps are executed in order:  
\begin{enumerate}
    \item 
    A random node is sampled from the configuration space \(\mathcal{X}\).  
    \item 
    The nearest node in the tree is identified, and the tree extends toward the sampled node within a specified step size.  
    \item 
    If the resulting path lies entirely within the free space \(\mathcal{X}_{free}\), the new node is added to the tree.  
    \item 
    To improve the path, RRT* performs a local optimization step where the newly added node is checked against nearby nodes in the tree. If a shorter path to the new node exists through one of these neighbors, the tree is restructured to reflect the improved path.  
    \item 
    This process of node extension and path optimization continues until the goal node is added to the tree and the path is sufficiently optimized.  
\end{enumerate}

Compared to using RRT as the base algorithm, RRT* minimizes the overall path cost and ensures that the tree converges to the optimal path over time, typically resulting in a more efficient and smoother trajectory. 
This optimization is achieved through the rewiring of the tree as the algorithm progresses, refining the paths between nodes. The main program is outlined in Algorithm \ref{Sample-Algorithm}, where the sampling step ($Sampler$) is implemented as Algorithm \ref{RRT-Sampler}.

\begin{algorithm2e}
  \SetAlgoLined
  \KwData{$\mathcal{X}, x_s, x_g$}
  \KwResult{$\mathcal{P}$}

  $x_s \rightarrow \mathcal{X}_{tree}$\;
  \While{$x_g$ is not in $\mathcal{X}_{tree}$}{
  
    \tcc{Update knowledge}
    
    $\Delta \mathcal{X}_{obs} \leftarrow f(Env(t), t, x_t)$\;
    
    $\mathcal{X}_{obs} \leftarrow \mathcal{X}_{obs} \cup \Delta \mathcal{X}_{obs}$\;
    
    \tcc{Take new sampled node}
    
    $x_{rand} \leftarrow Sampler(f, \mathcal{X}_{tree}, \mathcal{X})$\;
    
    $x_{nearest} \leftarrow Nearest(\mathcal{X}_{tree}, x_{rand})$\;
    
    $x_{next} \leftarrow Steer(x_{rand}, x_{nearest}, \epsilon)$\;

    \tcc{Check collsion}
    
    \If{CollsionFree}{
      $\mathcal{X}_{nearby} \leftarrow Nearby(\mathcal{X}_{tree}, x_{next})$\;
      
      $\mathcal{X}_{tree} \leftarrow \mathcal{X}_{tree}\cup \{x_{next}\}$\;
      
      $\mathcal{X}_{tree} \leftarrow Rewire(\mathcal{X}_{tree}, \mathcal{X}_{nearby}, x_{next})$
      }
    }
    
  $\mathcal{P} \leftarrow GetPath(\mathcal{X}_{tree}, x_s, x_g)$\;
  \caption{Base Algorithm}
  \label{Sample-Algorithm}
\end{algorithm2e}

\begin{algorithm2e}
  \SetAlgoLined
  \KwIn{$\mathcal{X}$}
  \KwOut{$x_{rand}$}
  
  $p \gets$ Probability of sampling goal node\;
  
  $r \gets$ Random number between 0 and 1\;
  
  \eIf {$r \leq p$}{
    $x_{rand} \gets x_g$\;
  }
  {$x_{rand} \leftarrow UniformSampling(\mathcal{X})$\;}
  
  \caption{RRT* Sampler}
  \label{RRT-Sampler}
\end{algorithm2e}

\subsection{Transformer-Based Sampler}
To implement the base RRT* algorithm, the key challenge lies in efficiently sampling states from the configuration space. Here, we propose a transformer-based sampler, as illustrated in Figure~\ref{Sampler_model}. Specifically, the sampler model consists of three main modules: the feature extractor, the transformer encoder, and the condition validator. The details of each component are as follows:

\subsubsection{Feature Extractor}
In the feature extractor part, we use convolutional layers to process the environment map. 
This module extracts useful information for sampling, such as obstacle locations and free space, from the original environmental data, which is then further processed by the transformer network.  

The original environment information is represented as a 2D or 3D cost map, which is processed through CNN (for 2D maps) or 3D-CNN (for 3D maps). We employ a \(3 \times 3\) convolution kernel with padding set to 1 to ensure that the spatial dimensions of the environment remain unchanged before and after convolution.  

After passing through three convolutional layers, the dimensionality of the feature space is transformed into \(d_{model}\), where \(d_{model}\) is the dimensionality of the transformer model. For example, in a 2D environment, the input to the CNN is a map of size \((1, \text{height}, \text{width})\), and the output is a feature map of size \((d_{model}, \text{height}, \text{width})\). This feature map captures essential spatial information about the environment, enabling the transformer network to make informed decisions during the sampling process.

\subsubsection{Transformer Encoder}
Once the environmental information is processed by the CNN, we further utilize a transformer model to process both the node positions and the processed environmental features. This ultimately predicts the next sampling node.

Specifically, the transformer is a network architecture composed of self-attention, multi-head attention, positional encoding, and fully connected layers. It is typically divided into two main components: the encoder and the decoder. Here, we only use the encoder part of the transformer to learn long-range dependencies between past nodes and the current environment, providing a more informed sampling node.

\paragraph{Input Processing. }We use the previous sampling nodes \(\mathcal{X}_{tree}\) as the original input to the transformer model. Since the number of previous sampling nodes varies each time, we first pad the sequence to a fixed maximum length. These nodes are then embedded into a higher-dimensional space using a simple linear transformation, uniformly mapping them to \(d_{model}\) dimensions. For example, in a 2D environment, we input a series of nodes with shape \((2, 1)\) and obtain embedded nodes of shape \((2, d_{model})\).  

\paragraph{Positional Encoding. }For the positional encoding of the transformer model, we take the features obtained by the Feature Extractor as input. These features are encoded using sine and cosine functions with different frequencies to effectively capture positional information. For instance, in a 2D environment, the encoding method is defined as:  
\begin{align}
PE(\text{x, y}, 2i) =& \sin\left(\frac{\text{x}}{10000^{\frac{2i}{d_{\text{model}}}}}\right),\\
PE(\text{x, y}, 2i+1) =& \cos\left(\frac{\text{y}}{10000^{\frac{2i}{d_{\text{model}}}}}\right),
\end{align}
where \(\text{x, y}\) represents the position, \(i\) is the dimension index, and \(d_{\text{model}}\) denotes the dimensionality of the model.  
 
The core of the transformer lies in the attention mechanism, whose primary computation is expressed as:  
\[
\text{Attention}(Q, K, V) = \text{softmax}\left(\frac{QK^\top}{\sqrt{d_k}}\right)V,
\]  
where \( Q \), \( K \), and \( V \) denote the query, key, and value vectors, respectively. These vectors are obtained by linear transformation of the vectors composed of the environment features and the previous nodes. The term \( d_k \) represents the dimensionality of the key vectors, and the softmax function ensures that the attention weights are normalized across all keys. Through the multi-head attention mechanism and the fully connected layer, we can get the final output, i.e., new sample node with shape $(d_{space}, 1)$, where $d_{space}$ is the dimension of the sampling space, which is 2 for 2D environment and 3 for 3D environment.

\subsubsection{Condition Validator}
Finally, the condition validator serves as the third component of the \textit{Sampler}, responsible for determining when the model should stop sampling. In transformer-based generative models, deciding when to halt the generation process is a critical consideration. For instance, in language models, this is typically managed by introducing a stop symbol or setting a maximum output length \cite{guo2024stop}.  
In our approach, the decision to continue sampling or stop and output the path is based on whether the newly sampled node is sufficiently close to the goal. If the sampled point is within a predefined threshold distance to the goal, the model terminates the sampling process and outputs the final path. This mechanism ensures that the algorithm efficiently converges to a solution while avoiding unnecessary computations.

\begin{algorithm2e}[t]
  \SetAlgoLined
  \KwIn{$\mathcal{X}_{tree}, Env(t)$}
  \KwOut{$x_{rand}$}
  $\alpha \gets$ Probability of uniform sampling\;
  
  $r \gets$ Random number between 0 and 1\;
  
  \eIf {$r \leq \alpha$}
  {$x_{rand} \leftarrow UniformSampling(\mathcal{X})$\;}
  {\tcc{Feature Extraction}

  $Env Feature \leftarrow Conv(Env)$\;

  \tcc{Model Sampling}
  
  $x_{rand} \leftarrow Transformer(\mathcal{X}_{tree}, Env Feature)$\;}
  
  \caption{Transformer Mixed Sampler}
  \label{Transformer-Sampler}
\end{algorithm2e}

With the above components detailed, we present the overall sampler as shown in Algorithm \ref{Transformer-Sampler}. Particularly, to ensure the probabilistic completeness of the sampling-based algorithm and introduce a certain degree of randomness, we incorporate a parameter \(\alpha\) that controls the balance between uniform and transformer-based sampling. Specifically, at each sampling step, the algorithm has a probability \(\alpha\) of performing uniform sampling and a probability \(1-\alpha\) of using transformer-based non-uniform sampling.   
This hybrid approach combines the strengths of both methods: the pure randomness of uniform sampling ensures exploration of the configuration space, while the transformer-based sampling leverages environmental and historical information to guide the search more efficiently. By adjusting \(\alpha\), the algorithm can balance exploration and exploitation, improving both the robustness and efficiency of the planning process.

\subsection{Training Details}
To train the neural network, we construct a training set consisting of 8,000 randomly generated environments, with the optimal paths obtained using the A* algorithm. 
The dataset is formatted as \((\mathcal{X}|(n), x_{new}(n+1), Env(n))\), where:  
\begin{itemize}
    \item 
    \(\mathcal{X}|(n)\) represents the first \(n\) nodes on the path generated by the A* algorithm,   
    \item   
    \(x_{new}(n+1)\) is the next node generated by the A* algorithm, and  
    \item 
    \(Env(n)\) is the environmental information at step \(n\).  
\end{itemize}
By training the model on these optimal path planning experiences, the transformer learns a more effective sampling strategy based on historical data, which is likely to lead to faster convergence to a feasible path.  

We use the Mean Squared Error (MSE) as the loss function, which calculates the error between the predicted next sampling node and the true target sampling node. The MSE is given by:  
\[
L_{MSE} = \frac{1}{N} \sum_{i=1}^{N} \left( x_{new}^i - \hat{x}_{new}^i \right)^2,
\]  
where \(x_{new}\) is the predicted node, \(\hat{x}_{new}\) is the actual target sampling node, and \(N\) is the batch size.  
During each training iteration, we use this loss function to perform backpropagation, simultaneously updating all the learnable parameters of both the Transformer and CNN.  

The benefit of updating both the Transformer and CNN parameters simultaneously during backpropagation is that it enables the entire model to learn more efficiently as a unified system. This approach allows the CNN to fine-tune the extraction of environmental features while the Transformer learns to incorporate these features for sampling and path prediction. By jointly optimizing both components, the model can adapt more effectively to the relationships between the environment and node positions, leading to improved performance in tasks like path planning. Additionally, this integrated training strategy ensures that the feature extraction and decision-making processes are aligned, enhancing the model's ability to generalize to new, unseen environments.

\section{Experimental Results}
In this section, we conduct a series of experiments to evaluate the performance of the proposed algorithm. First, we provide illustrative examples in both 2D and 3D environments, demonstrating how the randomization rate \(\alpha\) influences the algorithm's performance and highlighting its importance in balancing exploration and exploitation. Next, we compare our algorithm with existing RRT* and its variants on a large set of randomly generated environments to demonstrate its efficiency in terms of planning time and  path quality. Finally, we present a case study in a simulation environment using ROS   and Gazebo. 

\subsection{Experimental Settings} 
Our sampler model is implemented using PyTorch and trained on a single GPU (RTX 3090). 
The optimizer used is Adam \cite{kingma2014adam} with default parameters, and the learning rate is set to 0.0001, remaining consistent throughout the training process.
The transformer model is configured with the following parameters: \(d_{model} = 64\), \(n_{head} = 6\), and 6 encoder layers. 
For the 2D environment, the batch size is set to 256, and the model contains 0.52M parameters. 
For the 3D environment, the batch size is set to 128, and the model contains 0.56M parameters.  
 
We consider a scenario where the algorithm must find a safe path from a starting node to an end node in a randomly generated obstacle-filled environment.  Specifically, 
\begin{itemize}
    \item 
    For the 2D environment, we generate approximately 500 scenes with a map size of \(100 \times 100\). 
All obstacles are circular, with the number of obstacles per scene ranging from 16 to 20 and their radii varying from 0 to 12. Using circular obstacles ensures generality, even when obstacles overlap. 
The step size is set to 4, the rewire radius to 0, and the uniform sampling rate \(\alpha\) to 0.5.  
    \item 
For the 3D environment, we generate approximately 500 scenes with a map size of \(50 \times 50 \times 50\). 
All obstacles are spherical, with the number of obstacles per scene ranging from 6 to 10. 
The parameter configurations, such as step size, rewire radius, and uniform sampling rate, remain the same as in the 2D environment. 
\end{itemize}

\begin{figure}[t]
\centering
\includegraphics[scale=0.38]{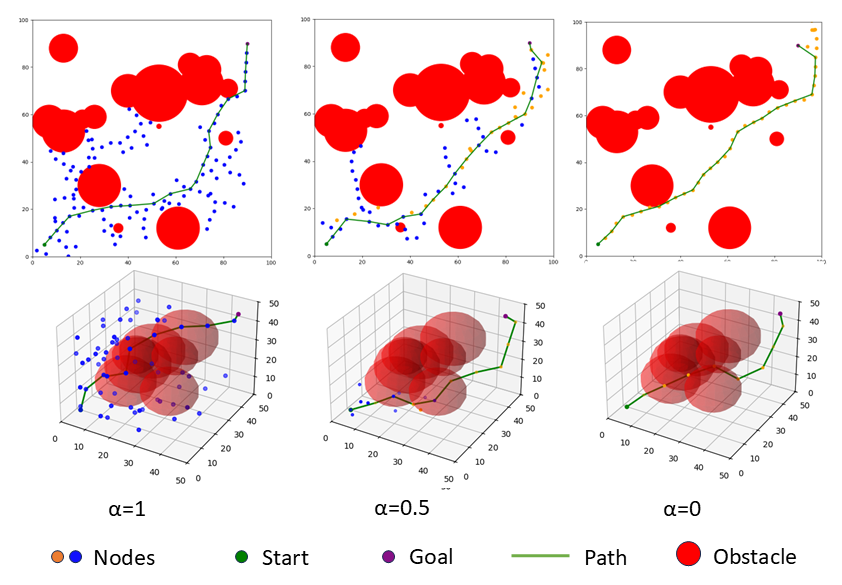}
\caption{Demonstration of the random tree generated under different randomization ratios \(\alpha\).}
\label{abation_study}
\end{figure} 

\begin{table}[t]
\centering
\renewcommand{\arraystretch}{1.5}
\caption{Statistical results under different randomization ratios \(\alpha\) across the 500 randomly generated 2D environments.}
\scalebox{1}{\begin{tabular}{|c|c|c|}
\hline
$\alpha$ & Average Nodes & Success Rate(\%) \\ \hline
0     &    47.02   &       84.14           \\ \hline
0.5   &    79.70   &        100.00          \\ \hline
1     &    275.14   &      100.00            \\ \hline
\end{tabular}
}
\label{abation_result}
\end{table}

\subsection{The Role of Hybrid Sampling}
Recall that our approach employs a hybrid sampling strategy by combining transformer-based sampling and uniform sampling with a ratio \(\alpha\). 
As illustrated in Figure \ref{abation_study}, when \(\alpha = 0\), 
the algorithm relies purely on transformer-based sampling, and when \(\alpha = 1\), it relies purely on uniform sampling. 
While transformer-based sampling often guides the trajectory toward the goal state more efficiently, requiring fewer sampling nodes, it may fail to achieve the planning task in some cases. Specifically, by focusing  on the current environment and prior nodes, the model may overlook certain regions, potentially resulting in fewer viable paths.

\begin{table*}[t]
\centering
\renewcommand{\arraystretch}{1.5}
\caption{Statistics on the comparison of our algorithm with other algorithms.}
\begin{tabular}{|c|c|c|c|c|c|c|}
\hline
\makebox[0.1\textwidth][c]{Scenario}     & \makebox[0.1\textwidth][c]{Method} & \makebox[0.1\textwidth][c]{Time(s)}  & \makebox[0.1\textwidth][c]{Nodes} & \makebox[0.1\textwidth][c]{Iterations} & \makebox[0.1\textwidth][c]{Initial Cost} & \makebox[0.1\textwidth][c]{Final Cost} \\ \hline
\multirow{4}{*}{2D Random Environment} & RRT*             & \textbf{0.15} &    286.04  &     335.19    &    159.00   & \textbf{128.67}       \\ \cline{2-7} 
                                       & NRRT*  & 0.80 &    202.45   &     314.91    &    151.53  & 132.47       \\ \cline{2-7} 
                                       & NIRRT*  & 1.37 &    164.56  &     211.96    &    147.13   & 132.10       \\ \cline{2-7} 
                                       & RRT*former             & 0.40 &    \textbf{79.70}    &    \textbf{159.22}    &    \textbf{139.45}  & 129.36       \\ \hline
\multirow{4}{*}{3D Random Environment} & RRT*   &      \textbf{0.04}         &    86.45   &     98.09     &     105.26   &    \textbf{77.71}       \\ \cline{2-7} 
                                       & NRRT*  &      0.72         &    35.99   &     54.19     &     98.49    &    79.16       \\ \cline{2-7} 
                                       & NIRRT* &      1.44         &    35.49   &     53.12     &     97.67    &    80.26       \\ \cline{2-7} 
                                       & RRT*former   &      0.29         &    \textbf{16.65}   &     \textbf{36.81}     &     \textbf{94.82}    &    78.27       \\ \hline
\end{tabular}
\label{result}
\end{table*}

\begin{figure*}[t]
	\centering
 	\begin{subfigure}[t]{0.24\linewidth}
		\centering
		\includegraphics[width=1\linewidth]{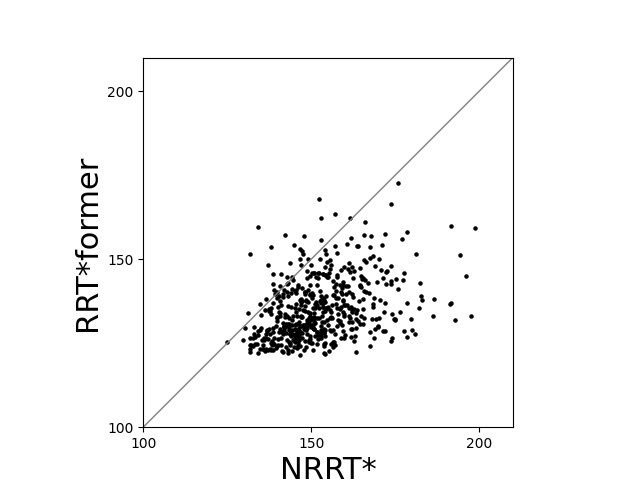}
		\caption{}
        \label{2d_init_path}
	\end{subfigure}	
	\begin{subfigure}[t]{0.24\linewidth}
		\centering
		\includegraphics[width=1\linewidth]{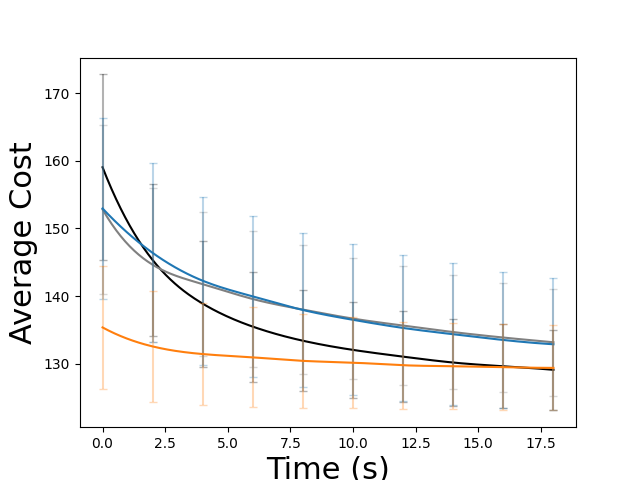}
		\caption{}
        \label{2d_optim_path}
	\end{subfigure}
	\begin{subfigure}[t]{0.24\linewidth}
		\centering
		\includegraphics[width=1\linewidth]{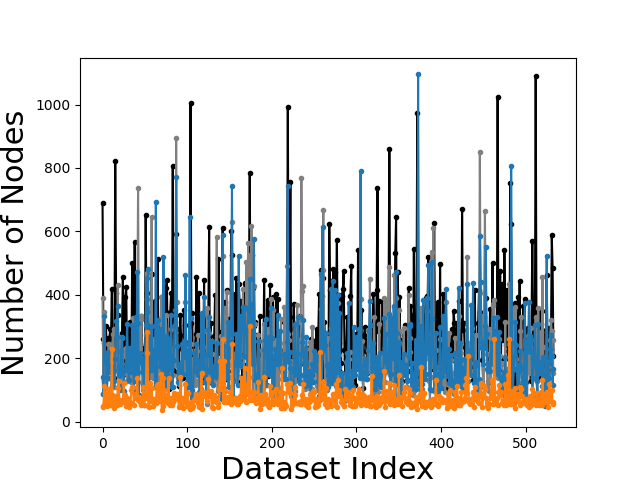}
		\caption{}
	\end{subfigure}
	\begin{subfigure}[t]{0.24\linewidth}
		\centering
		\includegraphics[width=1\linewidth]{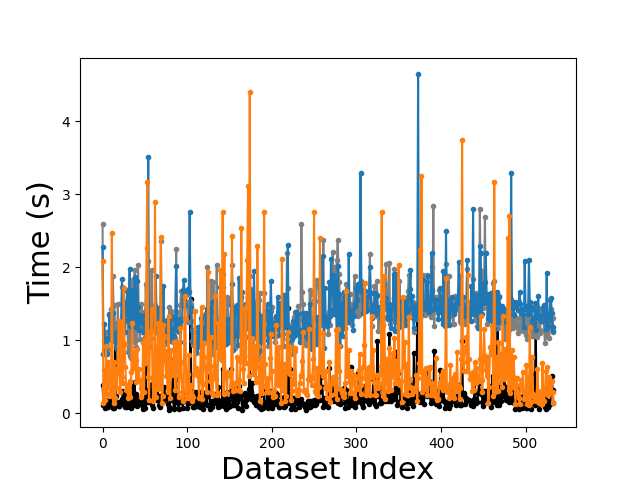}
		\caption{}
	\end{subfigure}

  	\begin{subfigure}[t]{0.24\linewidth}
		\centering
		\includegraphics[width=1\linewidth]{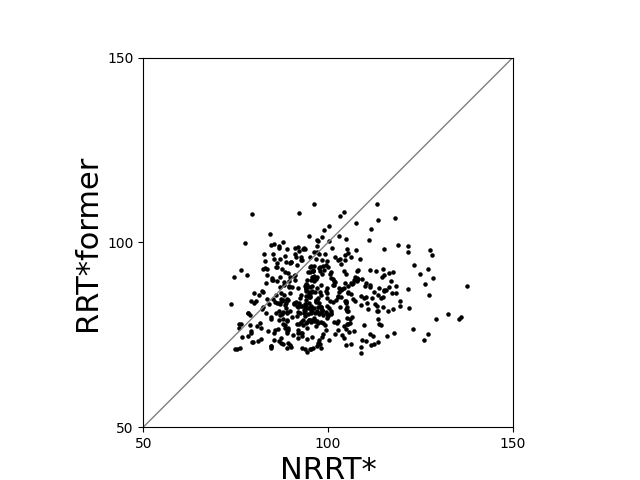}
		\caption{}
        \label{3d_init_path}
	\end{subfigure}
	\begin{subfigure}[t]{0.24\linewidth}
		\centering
		\includegraphics[width=1\linewidth]{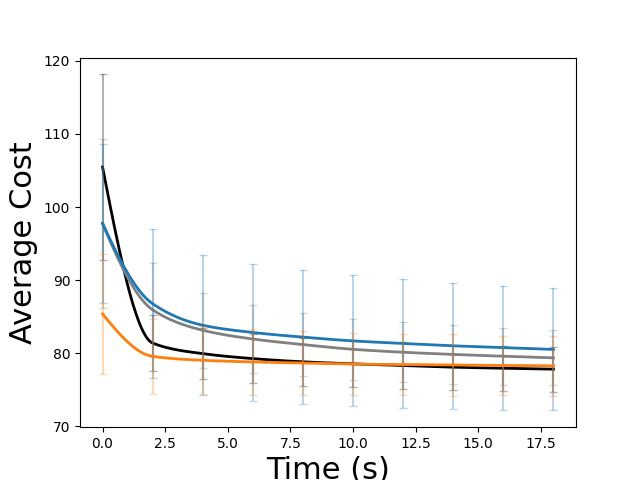}
		\caption{}
        \label{3d_optim_path}
	\end{subfigure}
	\begin{subfigure}[t]{0.24\linewidth}
		\centering
		\includegraphics[width=1\linewidth]{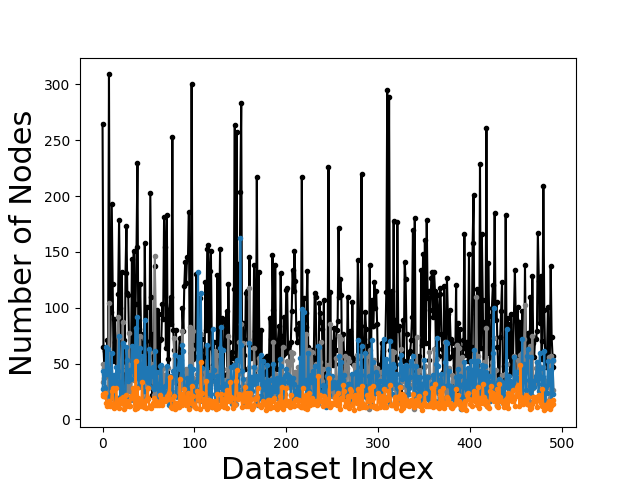}
		\caption{}
	\end{subfigure}
	\begin{subfigure}[t]{0.24\linewidth}
		\centering
		\includegraphics[width=1\linewidth]{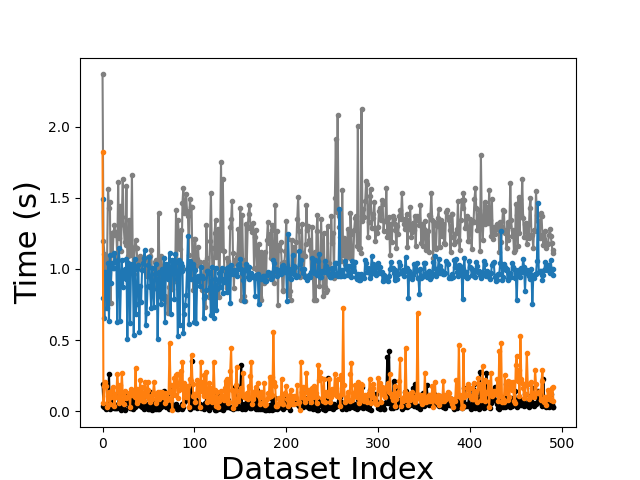}
		\caption{}
        \end{subfigure}

        \begin{subfigure}{0.7\linewidth}
		\centering
		\includegraphics[width=0.7\linewidth]{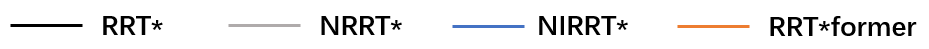}
	\end{subfigure}
	
 \caption{Figures (a)-(d) present the results for 2D environments, while Figures (e)-(h) present the results for 3D environments. Specifically:  
Figures (a) and (e) compare the initial path lengths found by NRRT* and RRT*former; 
Figures (b) and (f) compare the path optimization speed of each algorithm after finding the initial path;
Figures (c) and (g) compare the number of sampled nodes when each algorithm finds the initial path;
Figures (d) and (h) compare the time taken by each algorithm to find the initial path.}
 \label{Experiment_Detail}
\end{figure*}

To evaluate this more clearly, we conducted experiments on the randomly generated 500 scenes for the 2D environment case. We ran our RRT*former algorithm with parameters \(\alpha = 0\), \(\alpha = 0.5\), and \(\alpha = 1\), and computed the average number of nodes in the tree and the success rate of finding a safe path from the initial state to the goal. The results are shown in Table~\ref{abation_result}.  
The results indicate that while purely transformer-based sampling (\(\alpha = 0\)) generates fewer sampling nodes, it leads to a lower success rate due to its reduced randomness. On the other hand, purely uniform sampling (\(\alpha = 1\)) depends heavily on randomness and fails to incorporate environmental context or prior sampling information, resulting in a larger number of redundant nodes and inefficient exploration. In contrast, hybrid sampling (\(\alpha = 0.5\)) achieves a trade-off by leveraging existing information to reduce redundancy while preserving the randomness necessary for improving success rates.

\subsection{Comparison with Other Methods} 
To further verify the efficiency of our approach, we compare the proposed RRT*former algorithm with various existing algorithms that leverage neural networks to extract environmental features. Specifically, we compare  the following algorithms:
\begin{itemize}
    \item \textbf{RRT*} (Rapidly-exploring Random Tree*)
    \item \textbf{NRRT*} (Neural RRT*, \cite{wang2020neural})
    \item \textbf{NIRRT*} (Neural Informed RRT*, \cite{huang2024neural})
    \item \textbf{RRT*former} (Our Algorithm)
\end{itemize}
In NRRT* and NIRRT*,  we adopts the Pointnet \cite{qi2017pointnet} model  as provided in \cite{huang2024neural}. 

We run each algorithm on the randomly generated 500 2D environments and 500 3D environments. For each environment, we compute the following metrics:
\begin{itemize}
    \item \textbf{Time}: The average time required to find an initial path.  
    \item \textbf{Nodes}: The average number of nodes explored when finding the initial path.  
    \item \textbf{Iterations}: The average number of iterations needed to find the initial path.  
    \item \textbf{Initial Cost}: The average length of the path found initially (before optimization).  
    \item \textbf{Final Cost}: The average length of the path after 18 seconds of path optimization.  
\end{itemize}
The statistical metrics are shown in Table \ref{result}, and additional experimental details are provided in Figure~\ref{Experiment_Detail}. Based on the experiments, we draw the following observations:

\textbf{Time Efficiency: }  
One major advantages of RRT*former  is its substantial reduction in computation time. 
Compared to traditional methods like RRT*  and other learning-based algorithms such as  NRRT*  and NIRRT*, our method significantly shortens the time required to find an initial path. 
This improvement is primarily due to the model's reduced parameter complexity and its reliance on a smaller input of sampling node data, rather than processing the entire environment representation, such as images or point clouds. This streamlined input reduces both the computational burden and the overall path-finding time, making RRT*former more efficient.  

\textbf{Initial Path Cost:  }
Another significant advantage of RRT*former  is the quality of the initial path it generates. The initial cost of the path found by RRT*former  is consistently lower than that of other methods. This improvement stems from the transformer-based sampling strategy, which conditions its sampling process on both the current environmental state and the history of past sampling nodes. 

\textbf{Optimal Path and Final Cost:  }
The final cost after optimization is another area where RRT*former shows competitive performance. 
Although the final cost of RRT*former in both 2D and 3D environments is slightly higher than that of RRT*, the difference is minimal. 
Importantly, RRT*former achieves this result much faster than both traditional RRT* and learning-based methods. It can  converge to an optimal or near-optimal solution with significantly reduced computational cost.   

\textbf{Node Exploration Efficiency:  }
In terms of node exploration, RRT*former  demonstrates a significant advantage. 
It explores far fewer nodes to find a valid path compared to existing algorithms. 
This efficiency arises from the transformer-based sampling strategy, which learns to prioritize promising areas of the environment for sampling, rather than relying on blind random exploration.

\subsection{Simulation Deployments}
\begin{figure}[t]
	\centering
	\begin{subfigure}{0.325\linewidth}
		\centering
		\includegraphics[width=1\linewidth]{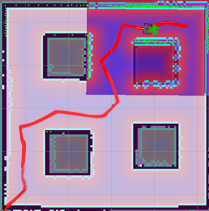}
	\end{subfigure}
	\centering
	\begin{subfigure}{0.325\linewidth}
		\centering
		\includegraphics[width=1\linewidth]{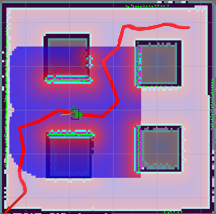}
	\end{subfigure}
	\centering
	\begin{subfigure}{0.325\linewidth}
		\centering
		\includegraphics[width=1\linewidth]{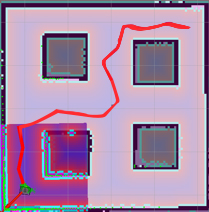}
	\end{subfigure}

 	\centering
	\begin{subfigure}{0.325\linewidth}
		\centering
		\includegraphics[width=1\linewidth]{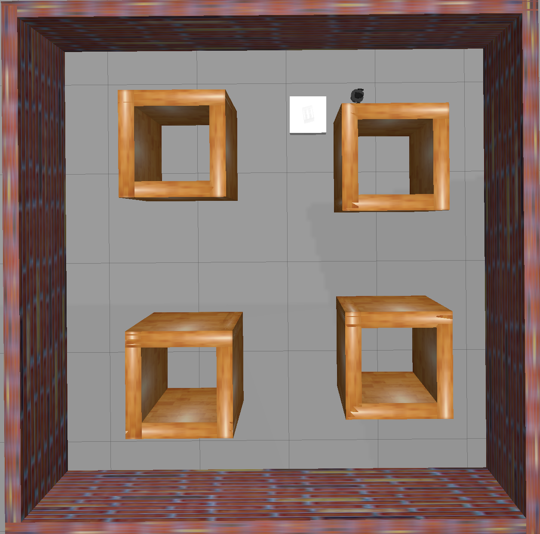}
		\caption{1s}
	\end{subfigure}
	\centering
	\begin{subfigure}{0.325\linewidth}
		\centering
		\includegraphics[width=1\linewidth]{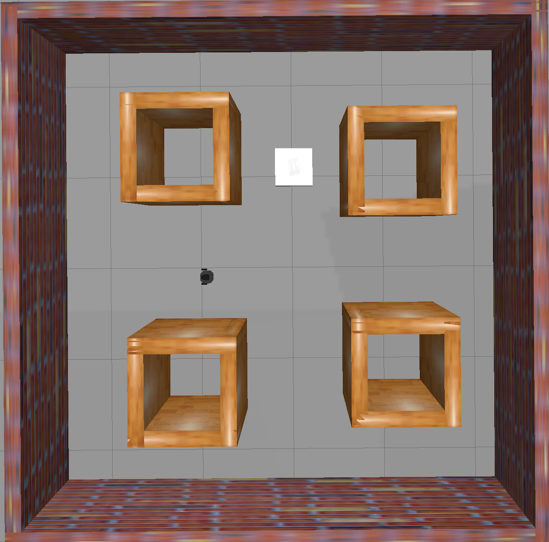}
		\caption{6s}
	\end{subfigure}
	\centering
	\begin{subfigure}{0.325\linewidth}
		\centering
		\includegraphics[width=1\linewidth]{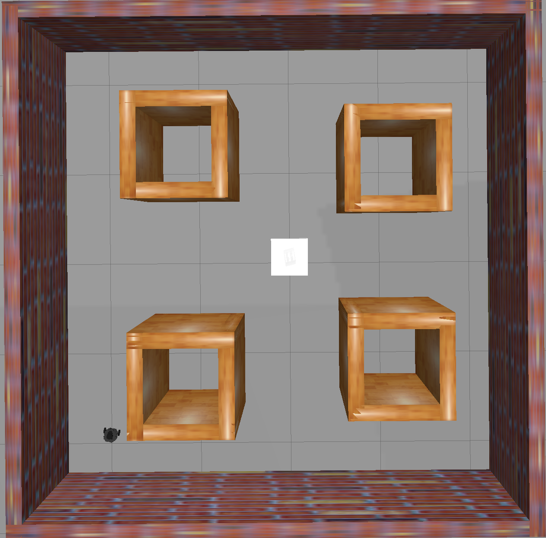}
		\caption{9s}
	\end{subfigure}
	\caption{Simulation Results in Gazebo:  
The three pictures above show the results of SLAM mapping, while the three pictures below depict the actual scenes in Gazebo. In the images, the four brown squares represent static obstacles, and the white square represents the moving obstacle. The red line indicates the path generated by our algorithm.}
	\label{simulation_demo}
\end{figure}

Finally, we deploy our algorithm and the model trained in a 2D random world to a TurtleBot3 Burger in the Gazebo simulator. As shown in Figure \ref{simulation_demo}, the map size is \(5\text{m} \times 5\text{m}\), with four static obstacles of size \(1\text{m} \times 1\text{m}\) and one moving obstacle (represented as the white square) of size \(0.4\text{m} \times 0.4\text{m}\). The robot's task is to move from the initial position \((-1\text{m}, 0\text{m})\) to the final position \((-5\text{m}, 3\text{m})\) while avoiding all obstacles. The robot can detect the moving obstacle within its sensing region, shown as the purple area.  

We map the \(5\text{m} \times 5\text{m}\) environment into a \(100 \times 100\) cost map in pixel space at a fixed ratio. After inputting the starting point, end point, and cost map into the model, we obtain a path in pixel space. This path is then reverse-mapped back to the actual simulation world and published to the robot for execution through ROS. Snapshots of the final simulation results are shown in Figure \ref{simulation_demo}, and the complete simulation video is available on our project website: \href{https://github.com/fengmingyang666/RRT-Net}.  

As demonstrated in Figure \ref{simulation_demo}, the robot successfully reaches the target state while avoiding all obstacles. Notably, since each sampling step is conditioned on both the current environmental context and the previously generated sampling nodes, the algorithm is inherently capable of avoiding dynamic obstacles in real-time. Unlike traditional random sampling methods, which treat each sampling event independently and without considering the evolving nature of the environment, this approach leverages past sampling information to generate more reasonable and forward-looking samples. 

\section{Conclusion}
In this paper, we propose a new sampling-based planning algorithm, RRT*former, which integrates the standard RRT* algorithm with the Transformer architecture. Our approach fully leverages the capabilities of the Transformer to process complex features from the environment map and learn context-dependent representations, such as the history of previously sampled states. Experimental results on randomly generated 2D and 3D environments demonstrate that, compared to existing algorithms based on RRT*, the proposed RRT*former offers considerable advantages in both path optimality and sampling efficiency.  
In future work, we plan to extend our approach to handle high-dimensional spaces and incorporate the kinematics of the robot to further explore the potential of this method.


\end{document}